\documentclass{article}

    \PassOptionsToPackage{numbers, compress}{natbib}
\usepackage[dblblindworkshop, final]{neurips_2026}
\workshoptitle{On-Device Intelligence: Foundation Models under Real-World Constraints}
\usepackage[T1]{fontenc}
\usepackage[utf8]{inputenc}
\usepackage{microtype}
\usepackage{graphicx}
\usepackage{subcaption}
\usepackage{placeins}
\usepackage{booktabs}
\usepackage{amsmath}
\usepackage{arydshln}
\usepackage{float}
\usepackage{hyperref}
\usepackage{url}
\usepackage{xurl}
\usepackage[most]{tcolorbox}
\usepackage[table]{xcolor}
\usepackage{kotex}
\usepackage{multirow}
\usepackage{wrapfig}
\usepackage{adjustbox}

\tcbset{
    safeexstyle/.style={
        enhanced, breakable,
        colback=green!4, colframe=green!45!black,
        boxrule=0.4pt, arc=2pt,
        left=6pt, right=6pt, top=4pt, bottom=4pt,
        fonttitle=\bfseries\footnotesize, coltitle=black,
        colbacktitle=green!18,
        attach boxed title to top left={xshift=4pt, yshift=-2pt},
        boxed title style={arc=2pt, boxrule=0.3pt, colframe=green!50!black},
    },
    artifactexstyle/.style={
        enhanced, breakable,
        colback=orange!5, colframe=orange!70!black,
        boxrule=0.4pt, arc=2pt,
        left=6pt, right=6pt, top=4pt, bottom=4pt,
        fonttitle=\bfseries\footnotesize, coltitle=black,
        colbacktitle=orange!20,
        attach boxed title to top left={xshift=4pt, yshift=-2pt},
        boxed title style={arc=2pt, boxrule=0.3pt, colframe=orange!70!black},
    },
}
\newtcolorbox{safeex}[1][]{safeexstyle, title=Input example, #1}
\newtcolorbox{artifactex}[1][]{artifactexstyle, title=Output example, #1}

\title{ixi-sLM: On-Device sLM for AI Call Agent Service}
\author{%
    Han-Sang~Lee \\ LG Uplus \And
    Kyuho~Lee \\ LG Uplus \And
    Juneyoung~Park \\ OptAI Inc. \And
    Seongbae~Lee \\ OptAI Inc. \AND
    Jin~Kyu~Park \\ LG Uplus \And
    Minjae~Kim \\ LG Uplus \And
    Yuri~Hong \\ OptAI Inc. \And
    Jinwoo~Lee \\ OptAI Inc. \And
    Youngwook~Kwon \\ OptAI Inc. \AND
    Seongwan~Kim \\ OptAI Inc. \And
    Seungheon~Hyeon \\ LG Uplus \And
    Jaeho~Lee$^{*}$ \\ OptAI Inc. \And
    Kiduk~Kwon$^{*}$ \\ LG Uplus
}

\begin{document}

\maketitle

{\renewcommand{\thefootnote}{*}%
\footnotetext{Corresponding authors: 
\texttt{jaeho.lee@opt-ai.kr}, 
\texttt{kwonkiduk@lguplus.co.kr}}}

\begin{abstract}
Running an AI call agent on-device makes it private, network-independent, and free of per-query cloud cost, but shipping one as a real-world service imposes constraints a cloud deployment never faces: $\sim$1\,GiB memory budget, in-call latency and thermal limits, static-integer NPU execution, Korean STT input far from sLM pretraining, and cloud-level quality expectations. We bring ixi-sLM, a 1.2B Korean sLM, to mobile NPUs through \emph{on-device modeling}: domain-adaptive modeling trims the vocabulary by call frequency to shrink the embedding and LM head that take nearly half the model under INT4 transformer weights and refits the tokenizer to call speech (18\% fewer tokens on call traffic), while task-aware quantization calibrates on real calls and assigns per-module mixed precision by activation sensitivity. A single domain-adapted model runs ixi-O's two production services, AI Call Summary and AI Call Answering, at 94\% of a state-of-the-art cloud LLM's task-weighted quality; our on-device modeling carries that quality onto the NPU with only a marginal drop in overall score, in an 803\,MiB model (28\% of FP16) at 23\% of the CPU's per-call energy. The same recipe deploys across Qualcomm and Apple NPUs and serves several million ixi-O subscribers in production.
\end{abstract}

\section{Introduction}
\label{sec:intro}

Privacy, network independence, and freedom from per-query cloud cost are pushing foundation-model intelligence onto the device \citep{mobilellm,lmmeter}; an AI call agent makes each pressure acute, as call content is sensitive, connectivity drops mid-call, and cloud cost scales with call volume.

Shipping such an agent as a real-world service, however, is governed by constraints a cloud deployment never faces. We identify five (\ref{sec:service}): \textbf{C1 (Memory)}, a $\sim$1\,GiB working-memory budget; \textbf{C2 (Latency and thermals)}, in-call responsiveness within a shared thermal envelope; \textbf{C3 (Execution model)}, precompiled static-integer NPU graphs committed at export time; \textbf{C4 (Input distribution)}, Korean STT traffic, rich in disfluencies and agglutinative morphology, that diverges from sLM pretraining and inflates token counts; and \textbf{C5 (Quality)}, direct user comparison against cloud services. No existing path satisfies all five at once: on-device frameworks fix tokenization and quantization as configuration rather than adapting them (C3, C4), and mobile quantization targets general benchmarks rather than service traffic (C4, C5).

\begin{figure}[t]
\centering
\adjustbox{trim=0 0.5cm 0 0,clip}{%
    \includegraphics[width=0.9\textwidth]{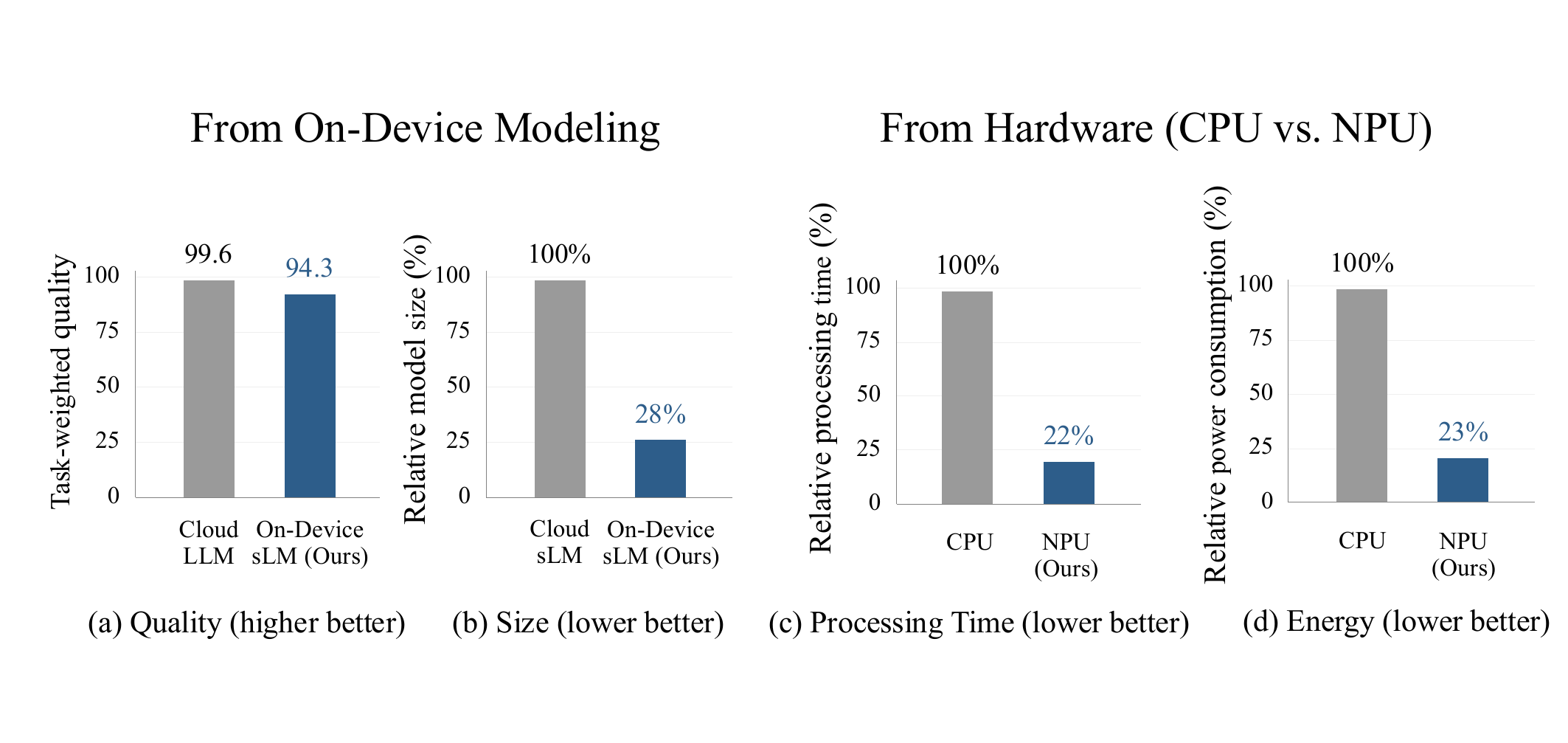}%
}
\caption{ixi-sLM (blue) compared with a cloud LLM, its FP16 counterpart, and the CPU: (a) 94\% of cloud quality, (b) 803\,MiB, 28\% of FP16, (c) 22\% of CPU latency, (d) 23\% of CPU energy.}
\label{fig:intro:result}
\end{figure}

We present ixi-sLM, the on-device model that powers ixi-O, an AI call agent used by several million subscribers. Every design decision in it follows from these constraints. Our contributions are: (i) \textbf{domain-adaptive modeling} that trims the vocabulary by call frequency and refits the tokenizer to Korean call speech, cutting 151 MiB from the deployed model and reducing call-traffic token counts by 18\% (C1, C4; \ref{sec:method:modeling}); (ii) \textbf{task-aware mixed-precision quantization} that calibrates on real calls and assigns per-module bit-width by activation sensitivity, with only a marginal drop in overall score (C3, C4; \ref{sec:method:quant}); and (iii) \textbf{a production deployment in ixi-O} that runs both services on one domain-adapted model, reaching 94\% of a SOTA cloud LLM's quality and largely retaining it through quantization, across Qualcomm and Apple NPUs (Figure~\ref{fig:intro:result}), evaluated against all five constraints (\ref{sec:exp}).

\section{Background and Related Work}
\label{sec:rel}

\paragraph{Generic on-device LLM Frameworks} such as llama.cpp \citep{llamacpp}, ExecuTorch \citep{executorch}, and ANEMLL \citep{anemll} execute LLMs on consumer hardware, from CPU/GPU to NPU-specific backends; all expose quantization and tokenization as fixed configuration, not as levers to adapt to a service's constraints (Appendix~\ref{sec:appendix:frameworks}, Table~\ref{tab:rel:comp}). 

\paragraph{Domain adaptation for sLMs} spans tokenizer-level vocabulary adaptation \citep{kim2024vocab} and knowledge-level domain-adaptive pretraining \citep{gururangan2020}; a domain-adapted model can outperform a larger general one on Telco tasks \citep{kim2025ixigen}. We use \emph{domain-adaptive modeling} for this service-facing adaptation to Korean call traffic. 

\paragraph{Quantization for on-device LLMs} is dominated by post-training quantization (PTQ), spanning weight-only quantization, activation/weight rebalancing, rotation- or trellis-based outlier suppression, and mobile-oriented quantization \citep{gptq,awq,smoothquant,omniquant,spinquant,qtip,mobilequant}; most assume GPU floating-point or per-token statistics that a static-integer NPU does not support well. We use \emph{hardware-aware optimization} for the quantization and export choices these constraints demand; together with domain-adaptive modeling it defines our \emph{on-device modeling}: a service-specific sLM both accurate for its call domain and executable within an NPU budget.
\section{ixi-O Service and Deployment Constraints}
\label{sec:service}

\begin{figure}[t]
\centering
\includegraphics[width=\textwidth]{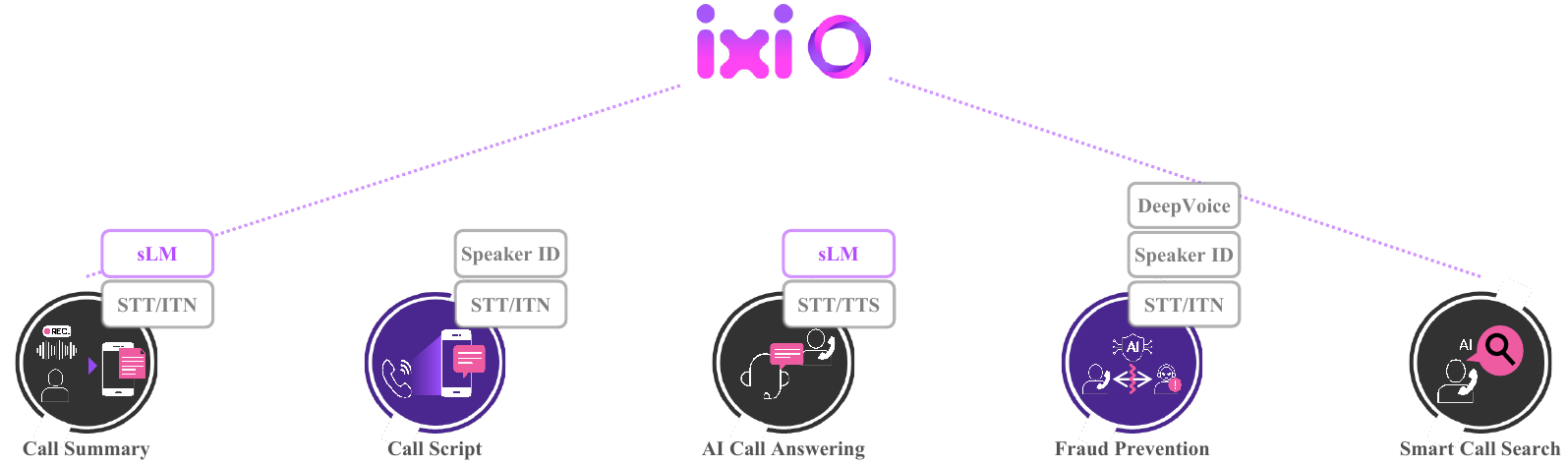}
\caption{Overview of ixi-O, our on-device AI call agent, and its core functions: AI call summary, real-time call script, AI call answering, fraud prevention, and smart call search.}
\label{fig:ixio_overview}
\end{figure}

\begin{figure}[h]
\centering
\includegraphics[width=\textwidth]{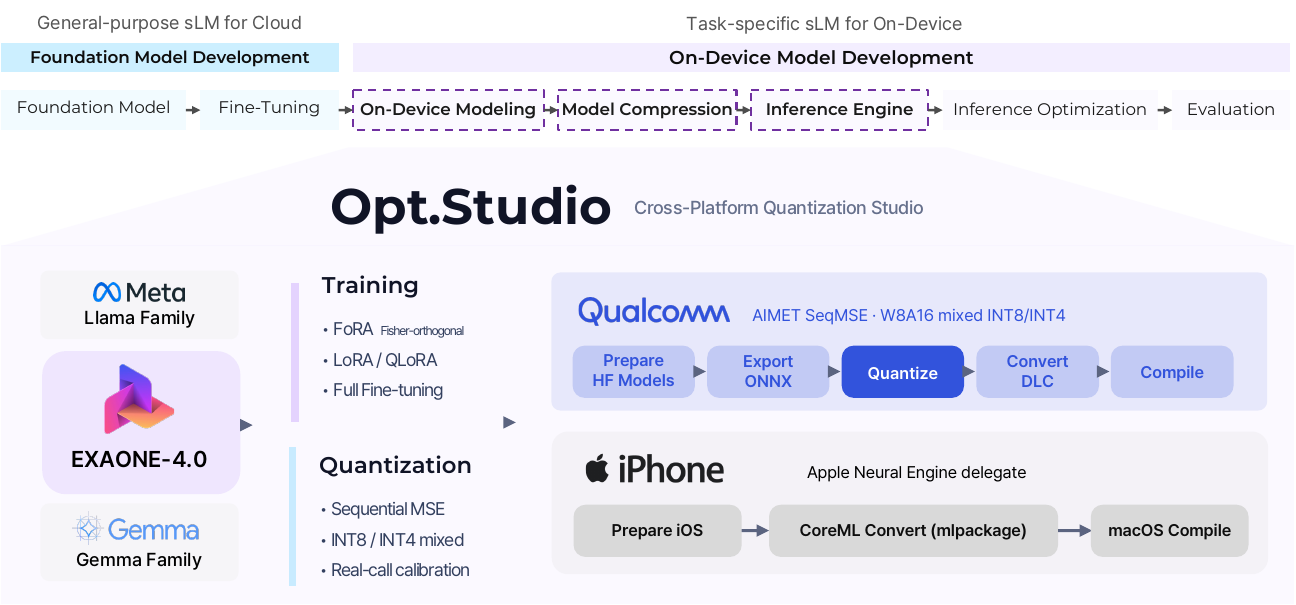}
\caption{Overview of the on-device sLM converting pipeline. Opt.Studio applies task-aware quantization and then exports the model to Qualcomm and Apple NPUs from a single recipe.}
\label{fig:process}
\end{figure}

\paragraph{Services and sLM tasks.}
Of the functions in the full ixi-O agent (Figure~\ref{fig:ixio_overview}, Appendix~\ref{sec:appendix:ixio_service}), two run on the sLM, each as a task configuration of one shared base model. \textbf{AI Call Summary} is triggered post-call: one inference pass turns the speech-to-text (STT) transcript into a \emph{one-line summary}, a \emph{detailed summary}, \emph{keyword extraction}, and \emph{task suggestion} for commitments the user made. \textbf{AI Call Answering} acts as an autonomous assistant when the user is unavailable, identifying the caller and their purpose and streaming that context to the user in real time.

\paragraph{Real-world deployment constraints.}
\label{sec:service:npuconstraints}
Turning these services into a shipped product imposes the five constraints of \ref{sec:intro} concretely. \textbf{C1:} devices span 6--12\,GiB of memory shared with the OS and other apps, budgeting $\sim$1\,GiB for the model (Appendix~\ref{sec:appendix:chiptier}). \textbf{C2:} summarization must complete within seconds of hang-up, answering must stream during a live call, and sustained CPU load throttles thermally. \textbf{C3:} mobile NPUs execute precompiled static-integer graphs (fixed shapes, calibration-precomputed activation ranges, a narrow operator and bit-width menu); a deployment commits at export time to a configuration the hardware executes natively, and any mismatch surfaces only on device. \textbf{C4:} inputs are Korean STT transcripts with recognition errors, fillers, and missing punctuation, far from sLM pretraining text. \textbf{C5:} output quality must stand comparison with the cloud services users already know.

\paragraph{Consequence: model selection.}
C1 and C2 rule out models above 3B on mobile \citep{mobilellm,llminflash}, pointing to a 1B-class model; C4 requires strength in Korean, where English-centric sLMs lose ground (Appendix~\ref{sec:appendix:chiptier}). We therefore build ixi-sLM on EXAONE4.0 1.2B \citep{exaone}, pretrained on large-scale Korean.

% \paragraph{Consequence: no generic path suffices.}ㄴㄴㄴㄴㄴ
% Before building our own pipeline, we tried to deploy the same Korean-native 1.2B sLM through existing NPU paths.\footnote{All generic on-device frameworks (llama.cpp, ExecuTorch, LiteRT, ANEMLL) were attempted in 2025-10.} LiteRT \citep{litert} failed to compile the model for NPU execution; ExecuTorch's QNN delegate exported and compiled it, and the graph ran correctly in PyTorch, yet on device its static-integer outputs were unusable, even though both frameworks run Llama-family models on the same hardware \citep{executorch}. C3 is thus not satisfiable by configuration alone, directly motivating \ref{sec:method:quant} and the framework below.
\paragraph{Consequence: no generic path suffices.}
Before building our own pipeline, we tried to deploy the same Korean-native 1.2B sLM through existing NPU paths. LiteRT \citep{litert} failed to compile the model for NPU execution; ExecuTorch's QNN delegate exported and compiled it, and the graph ran correctly in PyTorch, yet on device its static-integer outputs were unusable, even though both frameworks run Llama-family models on the same hardware \citep{executorch}. The ExecuTorch figures in Table~\ref{tab:app:ondevice_perf_full} therefore reflect speed only, not usable output. C3 is thus not satisfiable by configuration alone, directly motivating \ref{sec:method:quant} and the framework below.

\paragraph{Cross-platform framework.}
\label{sec:framework:oasis}
The model must be adapted to Korean call traffic (C4), quantized for static-integer NPUs (C3), and exported consistently across Qualcomm and Apple backends. We therefore built an in-house cross-platform studio carrying one trained sLM to both NPU targets from a single recipe (Figure ~\ref{fig:process}): it accepts EXAONE, Llama, or Gemma-family base models, applies domain training and task-aware quantization once, and exports each revision through the corresponding backend compiler, keeping both deployments numerically consistent.

\section{On-Device Modeling}
\label{sec:method}

This section describes how we build ixi-sLM to satisfy \ref{sec:service}'s constraints: domain-adaptive modeling (\ref{sec:method:modeling}) targets the input distribution (C4) and memory budget (C1); task-aware quantization (\ref{sec:method:quant}) targets the NPU execution model (C3) while preserving quality (C5).

\subsection{Domain-Adaptive Modeling}
\label{sec:method:modeling}

Domain-adaptive modeling comprises (i) \emph{domain training}, that is, continued training and task tuning on de-identified call corpora \citep{gururangan2020}, producing the FP16 \emph{Cloud sLM} of \ref{sec:exp} (Appendix~\ref{sec:appendix:domain_training}); and (ii) the tokenizer-level adaptation below, the footprint-relevant component we detail.

\paragraph{Vocabulary trimming.}

\begin{wraptable}{r}{0.55\linewidth}
\centering
\vspace{-1em}
\setlength{\tabcolsep}{5pt}
\captionsetup{skip=5pt}                 % 캡션과 표 사이 간격 확보
\caption{Merging accounts for the entire tokenization gain; trimming acts purely on footprint. Sizes in MiB.}
\label{tab:domain-adaptive-modeling}
\small                                  % 표 본문에만 작은 폰트 적용
\begin{tabular}{l r r r r}
\toprule
\textbf{Config.} & \textbf{Vocab.} & \textbf{Tok/call} & \textbf{$\Delta$} & \textbf{Size (MiB)} \\
\midrule
Original   & 102{,}400 & 1{,}324 & --        & 954 \\
Merge only & 102{,}400 & 1{,}087 & $-$17.9\% & 954 \\
Trim only  &  65{,}536 & 1{,}326 & $+$0.2\%  & 803 \\
Both       &  65{,}536 & 1{,}090 & $-$17.7\% & 803 \\
\bottomrule
\end{tabular}
\end{wraptable}

\begin{figure}[t]
\centering
\includegraphics[width=0.7\textwidth]{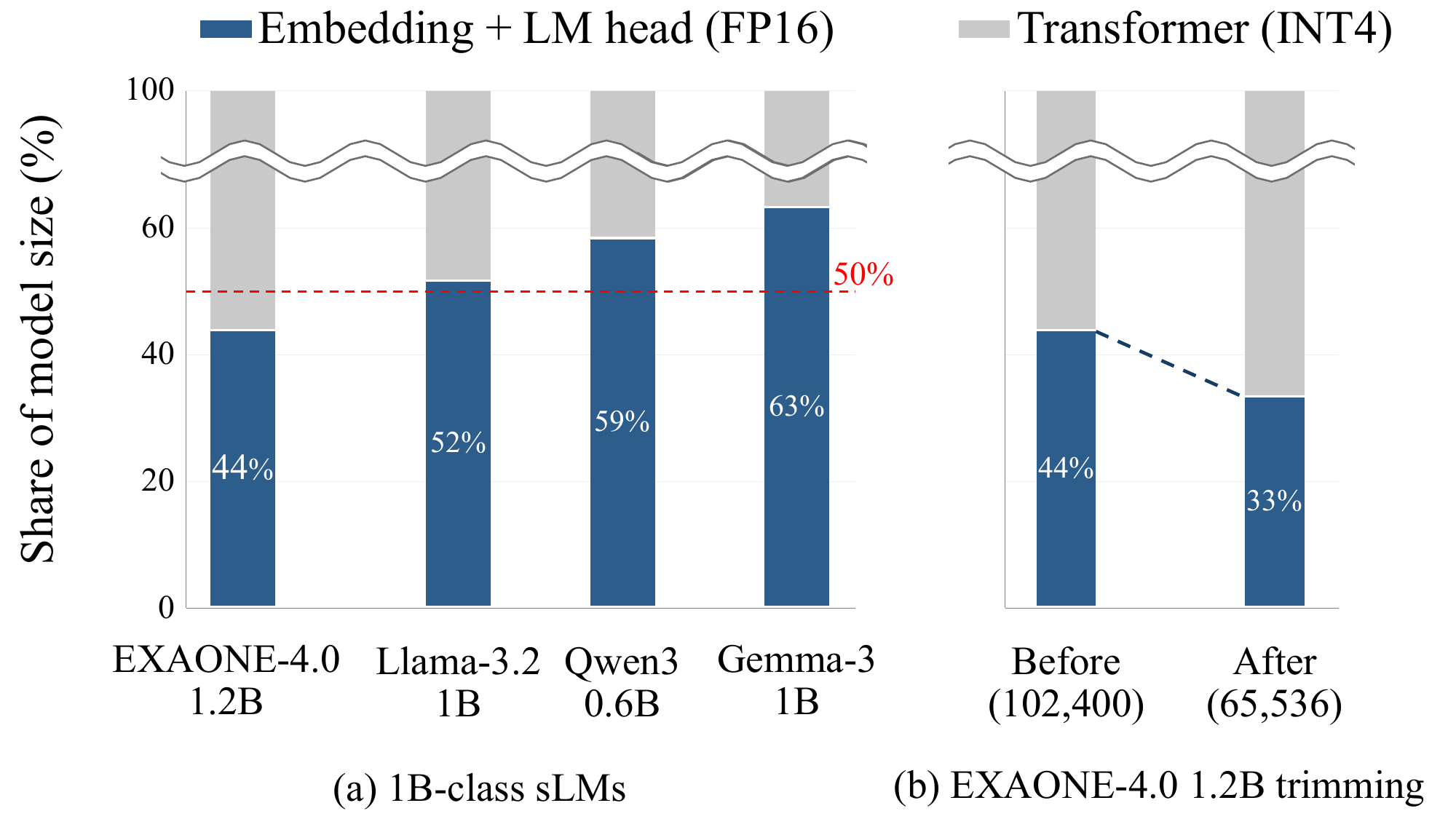}
\caption{(a) Embedding (FP16) vs.\ transformer (INT4) share across 1B-class sLMs. (b) shows the same FP16-embedding accounting as (a), isolating the effect of trimming; the deployed model quantizes the embedding further (4.2).}
\label{fig:vocabtrim}
\end{figure}

EXAONE4.0 1.2B carries a 102,400-token vocabulary; at hidden size 2,048 its tied embedding and LM head occupy 419\,MiB in FP16. Under the on-device configuration this class of model assumes, an FP16 embedding beside INT4 transformer weights, this single tensor accounts for 44\% of the model. The imbalance is structural: quantization shrinks the transformer weights but leaves the embedding in place, so its share grows as the rest contracts (Figure~\ref{fig:vocabtrim}a).

Vocabulary trimming attacks this footprint directly (C1) by removing tokens that never appear at inference: language-based trimming drops tokens outside the target languages, effectively lossless \citep{ushio2023}, and a frequency-based step removes rare tokens within them. The frequency cutoff is conventionally drawn over a general corpus, unsafe for a vocabulary already tuned to Korean: tokens rare in general text may be load-bearing in the domain. We instead redraw the cutoff over \emph{real call data}, where a token absent from call traffic is genuinely dead for the service, licensing a more aggressive cut. The two stages reduce the vocabulary from 102,400 to 65,536 tokens (36\%), lowering the tied embedding and LM head from 419\,MiB to 268\,MiB, the embedding share from 44\% to 33\% (Figure~\ref{fig:vocabtrim}b), and the total model from 954\,MiB to 803\,MiB (15.8\%).

\paragraph{Domain-aware tokenizer.}

EXAONE4.0 also ships with reserved slots its vocabulary never assigns; we reuse them to adapt the tokenizer by \emph{merging}, rather than removing, tokens, holding the vocabulary at 65,536 so the adaptation adds no footprint (C1, C4). On the input side, frequent Korean call expressions (greetings, acknowledgments, identified by frequency over call transcripts) collapse from several BPE tokens into one, shortening STT input; on the output side, fixed strings the service emits on every call (e.g., the section markers \texttt{"One Line Summary:"}, \texttt{"Detailed Summary:"}) are merged so a header costs a single decoding step.

The merges compound over a full call: across held-out calls, adaptation cuts token counts by 18\% (Table 1), and the reduction is what lets even the longest calls fit an entire transcript and its structured output in a single fixed window.

\subsection{Task-Aware Quantization}
\label{sec:method:quant}

Effective quantization for mobile NPUs must match both the input distribution (C4) and the hardware execution model (C3): PTQ estimates activation ranges from calibration data, so clean text such as WikiText can misrepresent STT-based call transcripts (recognition errors, fillers, missing punctuation), while global INT4 can over-compress sensitive modules. Our strategy combines real-call calibration, SeqMSE-based module-wise mixed precision, and top-k probability alignment for operating-point selection (details in Appendix~\ref{sec:appendix:quant_analysis}).

% \begin{table}[t]
% \centering
% \small
% \begin{tabular}{l c c cc}
% \toprule
% \multirow{2}{*}{\textbf{Quant.}} &
% \multirow{2}{*}{\textbf{Size (MiB)}} &
% \multirow{2}{*}{\textbf{Bits}} &
% \multicolumn{2}{c}{\textbf{ROUGE-L (\%)}} \\
% \cmidrule(lr){4-5}
% & & & \textbf{Call} & \textbf{WikiText} \\
% \midrule
% W4A16     & 726  & 4.10 & \textbf{82.23} & 79.42 \\
% \rowcolor{gray!12}MW4A16 & 803  & 5.09 & \textbf{80.63} & 79.47 \\
% W8A16     & 1111 & 8.10 & \textbf{84.29} & 81.67 \\
% \bottomrule
% \end{tabular}
% \caption{Real-call vs.\ WikiText calibration per quantization setting. Sizes are after vocabulary trimming; MW4A16 is the mixed-precision setting initialized from W8A16.}
% \label{tab:call_vs_wiki}
% \end{table}

\paragraph{Real-call calibration.}

\begin{wraptable}{r}{0.6\linewidth}
\centering
\vspace{-1em}
\captionsetup{skip=5pt}                 % 캡션과 표 사이 간격 확보
\caption{Real-call vs.\ WikiText calibration per quantization setting. Sizes are after vocabulary trimming; MW4A16 is the mixed-precision setting initialized from W8A16.}
\label{tab:call_vs_wiki}
% \small                                % 표 본문에만 작은 폰트 적용
\begin{tabular}{l r r rr}
\toprule
\multirow{2}{*}{\textbf{Quant.}} &
\multirow{2}{*}{\textbf{Size (MiB)}} &
\multirow{2}{*}{\textbf{Bits}} &
\multicolumn{2}{c}{\textbf{ROUGE-L (\%)}} \\
\cmidrule(lr){4-5}
& & & \textbf{Call} & \textbf{WikiText} \\
\midrule
W4A16     &  726  & 4.10 & \textbf{80.63} & 79.42 \\
\rowcolor{gray!12}MW4A16 &  803  & 5.09 & \textbf{82.23} & 79.47 \\
W8A16     & 1{,}111 & 8.10 & \textbf{84.29} & 81.67 \\
\bottomrule
\end{tabular}
\end{wraptable}

We compare WikiText and real-call calibration under W4A16 (4-bit weights, 16-bit activations), W8A16, and MW4A16 (W8A16 with low-sensitivity modules reduced to 4 bits via SeqMSE), evaluating with ROUGE-L \citep{rouge} between GPU (original) and NPU (quantized) outputs. Real-call calibration achieves higher ROUGE-L under all three settings (Table~\ref{tab:call_vs_wiki}), with the largest gain (+2.76) in MW4A16: task-aligned calibration preserves call-agent behavior more effectively than clean text. ROUGE-L here isolates the \emph{calibration-data} effect within each setting; the choice \emph{among} settings is made by distribution alignment and final task accuracy below.

\paragraph{Mixed-precision quantization.}
Modules in the same transformer layer respond very differently to low precision. Measuring each linear projection's sensitivity with Sequential MSE (SeqMSE), the activation difference between the original and quantized module on calibration data, we find \texttt{o\_proj} and \texttt{down\_proj} consistently far less sensitive than the other projections (Appendix~\ref{sec:appendix:quant_analysis}, Figure~\ref{fig:seqmse}); mixed precision therefore keeps high-SeqMSE modules at INT8 and lowers only low-SeqMSE modules to INT4.

\paragraph{Top-k probability alignment.}
To pick the operating point we compare per-layer top-k token distributions of the original and quantized models via the logit lens \citep{belrose2023tunedlens}. The 803\,MiB MW4A16 model preserves the final-layer alignment of the 1,111\,MiB W8A16 model, whereas global W4A16 loses alignment despite its competitive ROUGE-L in Table~\ref{tab:call_vs_wiki} (Figure~\ref{fig:alignment-tradeoff}); with the task accuracy of \ref{sec:exp:task_acc}, this fixes MW4A16 as the deployment point.

\section{Experiments}
\label{sec:exp}

\subsection{Experimental Setup}
\label{sec:exp:setup}
We evaluate against the constraints of \ref{sec:service}: quality (C5) via task accuracy, C1--C3 via on-device measurements, comparing \emph{Cloud LLM} (a frontier commercial LLM, the quality reference), \emph{Cloud sLM} (our domain-adapted EXAONE4.0 1.2B \citep{exaone} in FP16, \ref{sec:method:modeling}), and \emph{ixi-sLM} (the same model quantized, \ref{sec:method:quant}), on 100 de-identified held-out calls (up to 4K tokens) from the two ixi-O services, on a Galaxy S25 Ultra (Snapdragon 8 Elite) and an iPhone 17 Pro (A19 Pro), on CPU and NPU. Task accuracy is a weighted rubric score (0--100) graded by three frontier LLM judges under cross-evaluation \citep{llmjudge,geval} (rubric in Appendix~\ref{sec:appendix:acc}); on-device performance covers throughput, size, latency, and energy (Appendices~\ref{sec:appendix:ondevice_perf}, \ref{sec:appendix:power_thermal}), with latency, energy, and thermal results measured on the Samsung Galaxy S25 Ultra.

\subsection{Task Accuracy}
\label{sec:exp:task_acc}

\begin{wraptable}{r}{0.6\linewidth}
\centering
% \small
\vspace{-0.5em}
\caption{Task accuracy via LLM judge (weighted rubric, 0--100). Cloud sLM: our domain-adapted FP16 model; ixi-sLM: the same model quantized ($<$1\,GiB).}
\label{tab:exp:task_accuracy}
\begin{tabular}{l cccc}
\toprule
 & \textbf{Summary} & \textbf{Answering} \\
\midrule
Cloud LLM             & 99.60 & 99.00 \\
Cloud sLM             & 94.38 & 95.70 \\
\rowcolor{gray!12}ixi-sLM (Ours)  & 94.33 & 95.40 \\
\bottomrule
\end{tabular}
\end{wraptable}

Table~\ref{tab:exp:task_accuracy} separates two gaps toward C5. From the cloud LLM to our FP16 sLM, domain training (continued training and task tuning, Appendix ~\ref{sec:appendix:domain_training}) reaches 94.38 and 95.70, or 94.8\% and 96.7\% of the cloud LLM at a fraction of its size; from the FP16 model to ixi-sLM, quantization costs only 0.05 and 0.30 points, concentrated in task suggestion, a conditional sub-task evaluated on few calls (Appendix~\ref{sec:appendix:acc}). The drop is larger for schedule and location suggestion (99.55→87.05 and 97.61→83.95 in relative score), but the app only proposes these actions and the user confirms them (Figure~\ref{fig:app:ixioui}a).

\subsection{On-Device Performance}
\label{sec:exp:ondevice_perf}

\begin{wraptable}{r}{0.6\linewidth}
\centering
% \small
\vspace{-1em}
\caption{Per-call latency and energy, CPU vs.\ NPU.}
\label{tab:exp:ondevice_perf_summary}
\begin{tabular}{l cccc}
\toprule
 & \textbf{Latency (s)} & \textbf{Energy (mAh)} \\
\midrule
CPU (ExecuTorch)   & 45 & 7.79 \\
\rowcolor{gray!12}NPU (ixi-sLM) & 10 & 1.79 \\
\bottomrule
\end{tabular}
\end{wraptable}

Our on-device modeling shrinks the model from its 2.9 GiB FP16 checkpoint to 803 MiB (28\%) and cuts token counts by 18\%, meeting the C1 budget while fitting a full call in a single fixed window; the NPU cuts latency to 22\% (4.5$\times$) and per-call energy to 23\% (Table~\ref{tab:exp:ondevice_perf_summary}), and over a 300\,s sustained run stays below 35\,\textdegree C in surface temperature, beneath the 36\,\textdegree C throttling threshold, while the CPU crosses that threshold within 60\,s and plateaus near 45\,\textdegree C, meeting C2. The saving concentrates in compute-bound prefill ($\sim$17$\times$ less energy) versus memory-bound decode ($\sim$2$\times$; Appendix~\ref{sec:appendix:power_thermal}). Against other Snapdragon-NPU runtimes (Appendix~\ref{sec:appendix:ondevice_perf}), our engine is $\sim$1.5$\times$/1.7$\times$ faster than ExecuTorch on prefill/decode, over 30× faster than llama.cpp on prefill, in a 22\% smaller model than ExecuTorch.
\section{Conclusion}
\label{sec:conclusion}

We presented ixi-sLM, the on-device model behind the ixi-O call agent, built by treating real-world deployment constraints (C1--C5) as the starting point of the design: domain-adaptive modeling and task-aware quantization follow from them, and one 1.2B model, at 94\% of a SOTA cloud LLM's quality, runs across Qualcomm and Apple NPUs with only a marginal drop in overall score. Future work: inference optimization, on-device personalization, broader evaluation (Appendix~\ref{sec:appendix:limitations}).

\bibliographystyle{plainnat}
\bibliography{custom}
\newpage
\appendix
% \section{Appendix}
% The main paper is self-contained; this appendix provides supporting material: the framework comparison behind \ref{sec:rel} (\ref{sec:appendix:frameworks}), the hardware envelope and base-model selection behind \ref{sec:service} (\ref{sec:appendix:chiptier}), the module-level quantization diagnostics behind \ref{sec:method:quant} (\ref{sec:appendix:quant_analysis}), the full task-accuracy rubric (\ref{sec:appendix:acc}), complete on-device speed/size (\ref{sec:appendix:ondevice_perf}) and power/thermal (\ref{sec:appendix:power_thermal}) measurements, service-level context and prompt formats (\ref{sec:appendix:ixio_service}, \ref{sec:appendix:ixio_prompt}), and limitations (\ref{sec:appendix:limitations}).

\section{Generic On-Device Framework Comparison}
\label{sec:appendix:frameworks}

llama.cpp \citep{llamacpp} runs GGUF-quantized models on CPU and GPU and exposes an experimental Snapdragon NPU that executes but delivers an order of magnitude lower prefill throughput than delegating runtimes; it provides no NPU path for our Apple target. ExecuTorch \citep{executorch} compiles PyTorch models to mobile backends with partial NPU delegation via QNN, and ANEMLL \citep{anemll} compiles exclusively for the Apple Neural Engine. All three treat quantization and tokenization as fixed configuration chosen from a menu, rather than as levers to adapt to a service's input distribution and hardware target; this gap motivates our platform (\ref{sec:framework:oasis}).

\begin{table}[H]
\centering
%\small
\caption{Comparison of generic on-device LLM frameworks. ``Modeling'' denotes whether the framework exposes tokenizer/quantization adaptation as first-class levers ($\bigcirc$: supported, $\triangle$: partial, $\times$: unsupported).}
\label{tab:rel:comp}
\vspace{1em}
\begin{tabular}{l c c ccc}
\toprule
& & & \multicolumn{3}{c}{\textbf{Hardware}} \\
\cmidrule(lr){4-6}
\textbf{Framework} & \textbf{OS} & \textbf{Modeling} & \textbf{CPU} & \textbf{GPU} & \textbf{NPU} \\
\midrule
ExecuTorch & AOS/iOS & $\triangle$ & $\bigcirc$ & $\bigcirc$ & $\triangle$ \\
llama.cpp  & AOS/iOS & $\times$    & $\bigcirc$ & $\bigcirc$ & $\triangle$ \\
ANEMLL     & iOS     & $\times$    & $\bigcirc$ & $\times$   & $\bigcirc$  \\
\bottomrule
\end{tabular}
\end{table}

% \begin{figure}[H]
% \centering
% \begin{subfigure}{\textwidth}
%     \centering
%     \includegraphics[width=\textwidth]{figures/fig_4process.pdf}
%     \caption{On-device sLM development process}
%     \label{fig:framework_overview}
% \end{subfigure}
% \vspace{0.6em}
% \begin{subfigure}{\textwidth}
%     \centering
%     \includegraphics[width=\textwidth]{figures/fig_4architecture.pdf}
%     \caption{Cross Platform Quantization Studio}
%     \label{fig:framework_oasis}
% \end{subfigure}
% \caption{Overview of the Cross Platform Quantization Studio on-device sLM development pipeline. The top flow places our scope within the transition from a cloud foundation sLM to a task-specific on-device sLM, covering domain-adaptive modeling, model compression, and inference-engine export. the platform takes EXAONE, Llama, or Gemma-family models, applies training and task-aware quantization once, and exports the resulting model to Qualcomm and Apple NPU backends from a shared recipe.}
% \label{fig:framework_full}
% \end{figure}

\section{Domain Training}
\label{sec:appendix:domain_training}

The \emph{Cloud sLM} of \ref{sec:exp} is produced by domain training the EXAONE4.0 1.2B base model on de-identified call-domain corpora before any tokenizer or quantization step: continued training on call-domain text aligns the model with STT-style input (disfluencies, fillers, missing punctuation), and task-specific tuning on the two service formats (AI Call Summary and AI Call Answering) teaches the structured output schemas shown in Appendix~\ref{sec:appendix:ixio_prompt}. Training follows standard domain-adaptive practice \citep{gururangan2020}; all corpora are de-identified before use, and evaluation calls (\ref{sec:exp:setup}) are held out from training.
% TODO(authors): add corpus scale, training schedule, and hyperparameters if disclosable.

\section{Device Chip Tiers and Base-Model Selection}
\label{sec:appendix:chiptier}

\subsection{Mobile hardware envelope}
Target devices span 6--12\,GiB of system memory and NPUs delivering roughly 20--50 TOPS across product tiers. A deterministic on-device deployment must coexist with the OS and other applications, so we budget approximately 1\,GiB of working memory for the model. Combined with the latency and thermal headroom an active call requires, which rules out models above 3B, this envelope points to a 1B-class model.

\subsection{Korean capability of English-centric sLMs}
Within the 1B-class budget, the base model must also be strong in Korean. English-centric sLMs, including LLaMA3.2 1B, Gemma3 1B, and Qwen2.5 1.5B, score competitively on general English benchmarks but lose ground in Korean: their generic BPE vocabularies fragment Korean's agglutinative morphology into long token sequences, inflating the context window and degrading quality under a fixed budget. EXAONE4.0 1.2B, pretrained on large-scale Korean, tokenizes the same text more compactly and serves as our base model.

\clearpage
\section{Quantization Analysis}
\label{sec:appendix:quant_analysis}

This section provides the module-level evidence used to choose the MW4A16 mixed-precision setting in \ref{sec:method:quant}. We use two diagnostics. First, SeqMSE measures how much each projection module's activation changes when it is quantized, using the same real-call calibration distribution used for PTQ. Second, top-k probability alignment measures whether a quantized model preserves the token distribution of the original model across late layers and at the final prediction.

\begin{figure}[H]
\centering
\includegraphics[width=0.8\textwidth]{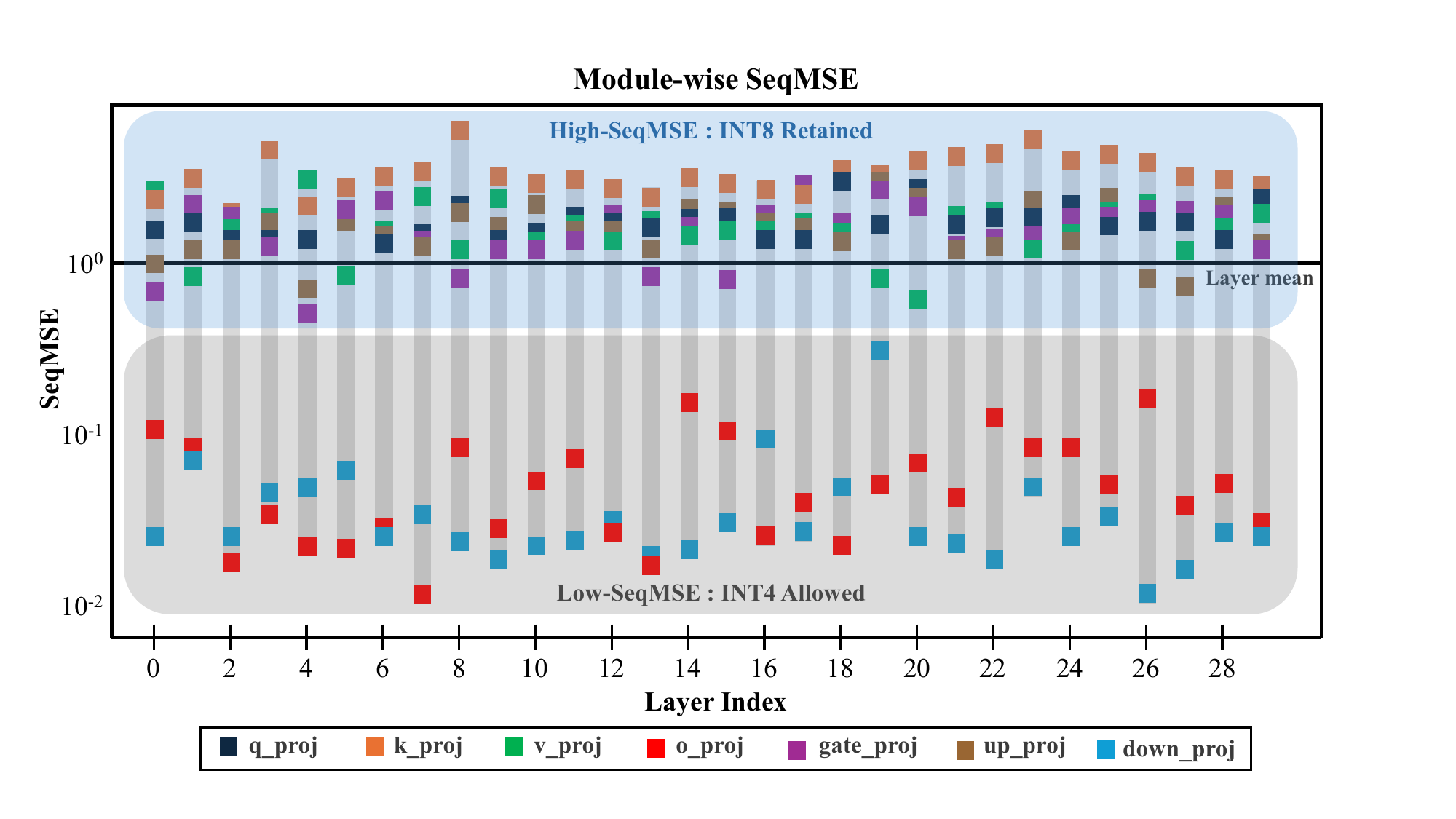}
\caption{Module-wise SeqMSE on EXAONE4.0 1.2B (W8A16). Modules above the threshold (blue region) are kept at INT8, while those below it (gray region) are lowered to INT4.}
\label{fig:seqmse}
\end{figure}

\begin{figure}[H]
    \centering
    \begin{minipage}{0.48\textwidth}
        \centering
        \includegraphics[width=\linewidth]{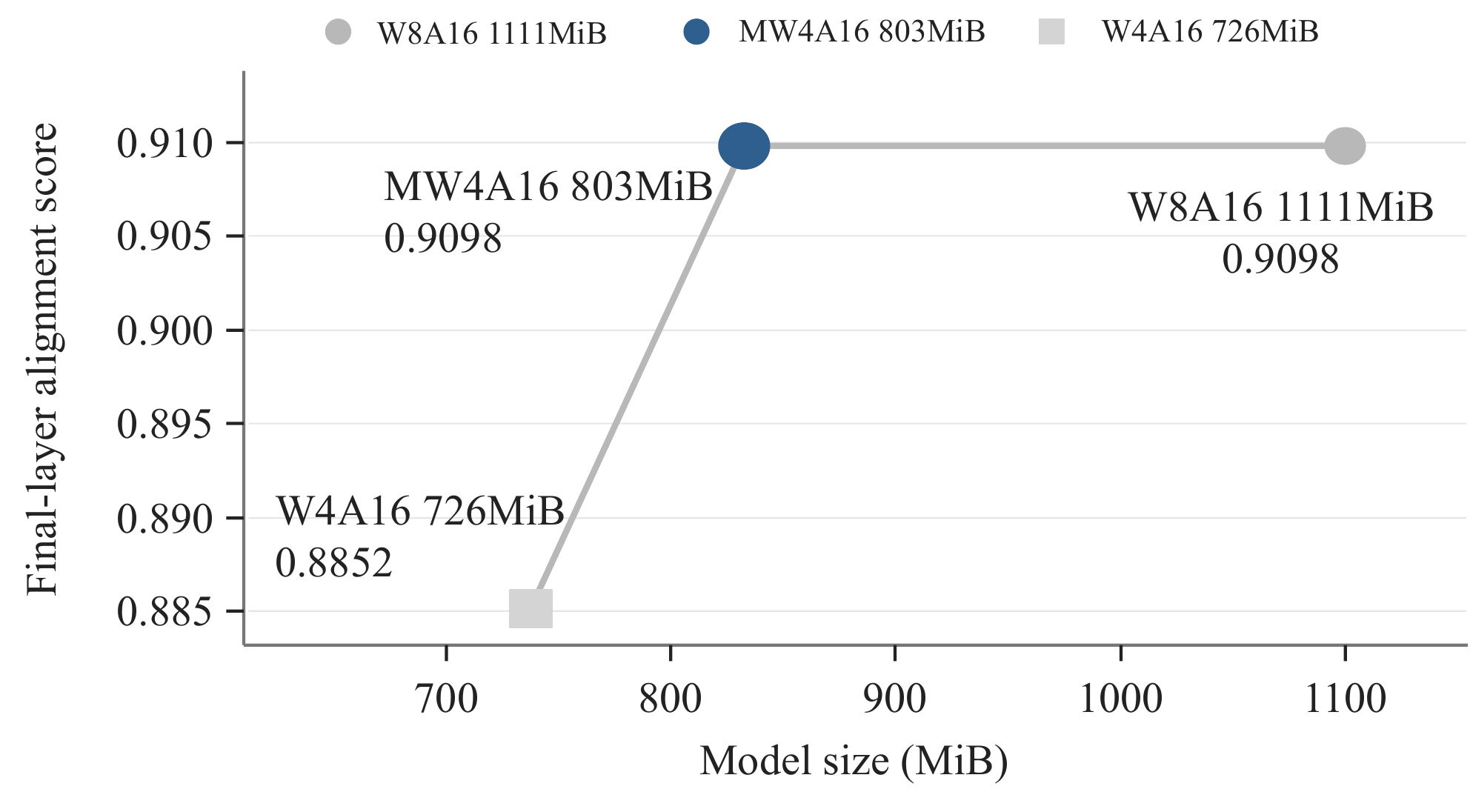}
        \small (a) Final-layer alignment score vs.\ model size.
    \end{minipage}
    \hfill
    \begin{minipage}{0.48\textwidth}
        \centering
        \includegraphics[width=\linewidth]{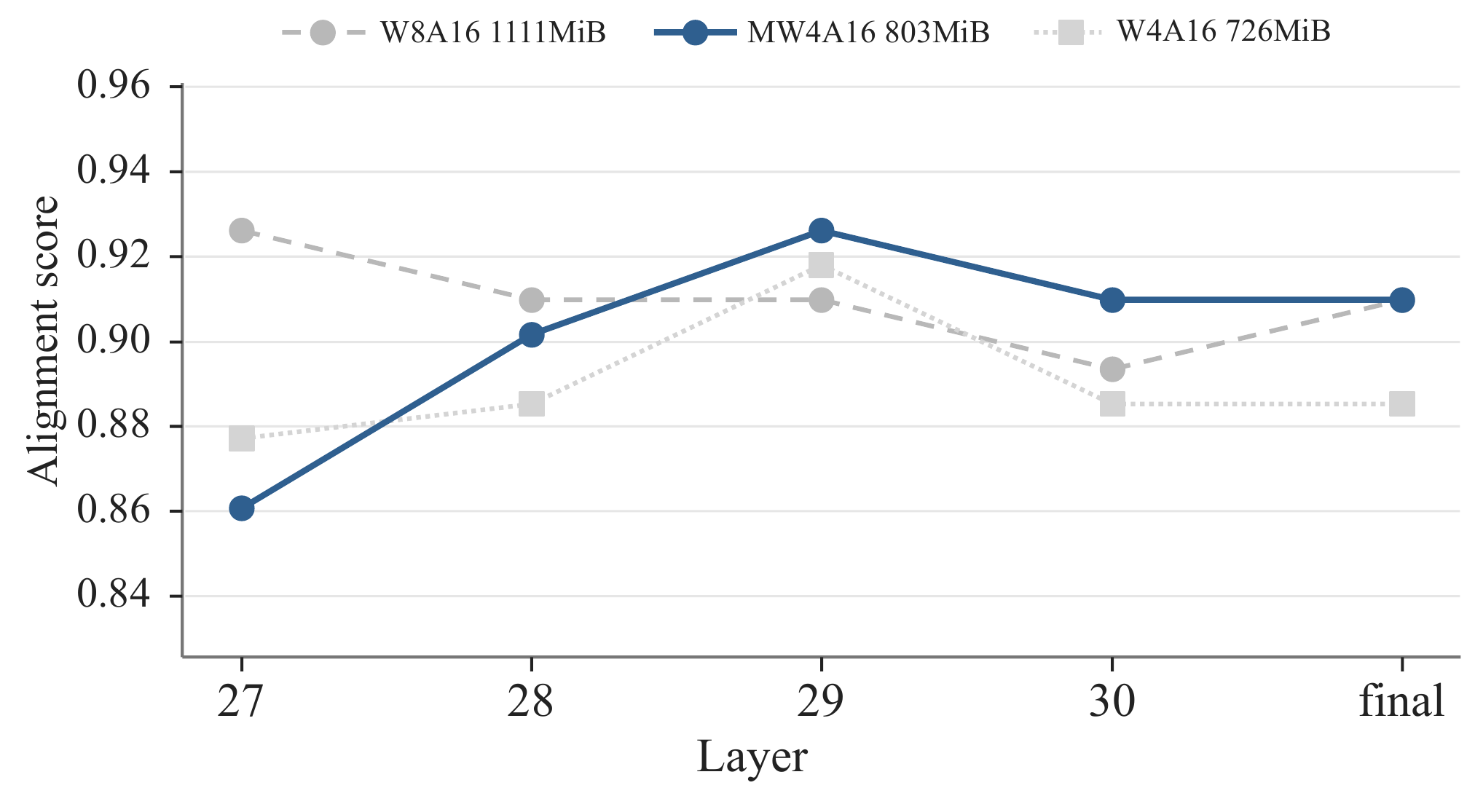}
        \small (b) Layer-wise alignment score, layer 27 to final.
    \end{minipage}
    \caption{Alignment score comparison across quantized model sizes. (a) shows the final-layer alignment score as a function of model size, where the 803\,MiB model preserves the final-layer alignment of the 1{,}111\,MiB W8A16 model at a smaller size. (b) shows the layer-wise alignment score in the late layers, indicating how each quantized model follows the original model before the final prediction.}
    \label{fig:alignment-tradeoff}
\end{figure}

% Figure~\ref{fig:seqmse} shows that quantization sensitivity is not uniform within a transformer block. The \texttt{o\_proj} and \texttt{down\_proj} modules consistently have lower SeqMSE than the other projections, so MW4A16 lowers these low-sensitivity modules to INT4 while retaining high-sensitivity modules at INT8. This module-wise assignment gives a smaller model than W8A16 without applying the same aggressive compression to every projection.

% Figure~\ref{fig:alignment-tradeoff} compares the resulting model-size operating points. The MW4A16 model preserves the final-layer alignment of the larger W8A16 model while reducing the footprint to 803\,MiB, whereas the smaller W4A16 setting loses alignment. This supports MW4A16 as the deployment point used in the main experiments: it stays within the on-device memory budget while preserving the original model's prediction behavior more closely than global 4-bit quantization.
Figures~\ref{fig:seqmse} and \ref{fig:alignment-tradeoff} motivate MW4A16 as the deployment setting: because the \texttt{o\_proj} and \texttt{down\_proj} modules consistently exhibit lower SeqMSE, quantizing only these low-sensitivity modules to INT4 reduces the model footprint to 803\,MiB while retaining the final-layer alignment of W8A16; in contrast, globally applying INT4 in W4A16 degrades alignment. Thus, MW4A16 satisfies the on-device memory budget while preserving the original model’s prediction behavior more closely than global 4-bit quantization.

\clearpage
\section{Task Accuracy: Full Rubric}
\label{sec:appendix:acc}
\begin{table}[H]
\centering
\caption{Detailed task-accuracy rubric for both services. For each sub-task we report the raw judge score (Raw), its relative score against a per-sub-task reference, and the weighted contribution to the total (Fin. = Wt × Rel.). The reference is the best score observed across systems and judges, so it can exceed the highest value shown. Raw scales differ across sub-tasks. Scores are averaged over a three-LLM judge panel under cross-evaluation.}
\label{tab:app:task_accuracy_full}
\vspace{1em}
\scriptsize
\setlength{\tabcolsep}{3pt}
\renewcommand{\arraystretch}{1.05}
\resizebox{\textwidth}{!}{
\begin{tabular}{l c rrr rrr rrr}
\toprule
& & \multicolumn{3}{c}{\textbf{Cloud LLM}}
  & \multicolumn{3}{c}{\textbf{Cloud sLM}}
  & \multicolumn{3}{c}{\textbf{ixi-sLM}} \\
\cmidrule(lr){3-5}\cmidrule(lr){6-8}\cmidrule(lr){9-11}
\textbf{Sub-task} & \textbf{Wt.}
& Raw & Rel. & Fin. & Raw & Rel. & Fin. & Raw & Rel. & Fin. \\
\midrule
\multicolumn{11}{l}{\textit{Service 1: AI Call Summarization}} \\
\midrule
One-line summary          & 25\% & 4.77  & 100.00 & 25.00 & 4.61  & 96.65 & 24.16 & 4.66  & 97.69 & 24.42 \\
Detailed summary          & 50\% & 22.00 & 100.00 & 50.00 & 21.42 & 97.36 & 48.68 & 21.48 & 97.64 & 48.82 \\
Keyword extraction        & 10\% & 4.81  & 100.00 & 10.00 & 4.67  & 97.09 & 9.71  & 4.73  & 98.34 & 9.83  \\
Task suggestion (schedule)& 6\%  & 4.22  & 95.91  & 5.75  & 4.38  & 99.55 & 5.97  & 3.83  & 87.05 & 5.22  \\
Task suggestion (location)& 6\%  & 4.49  & 97.40  & 5.84  & 4.50  & 97.61 & 5.86  & 3.87  & 83.95 & 5.04  \\
Task suggestion (contact) & 3\%  & 3.00  & 100.00 & 3.00  & 0.00  & 0.00  & 0.00  & 1.00  & 33.33 & 1.00  \\
\cmidrule(lr){1-11}
\textit{Total}            & 100\% &      &        & \textbf{99.60} &     &       & \textbf{94.38} &     &       & \textbf{94.33} \\
\midrule
\multicolumn{11}{l}{\textit{Service 2: AI Call Answering}} \\
\midrule
Intent classification     & 30\% & 5.00 & 100.00 & 30.00 & 4.86 & 97.20 & 29.16 & 4.85 & 97.00 & 29.10 \\
Information extraction    & 30\% & 4.90 & 98.00  & 29.40 & 4.81 & 96.20 & 28.86 & 4.79 & 95.80 & 28.74 \\
Guardrail                 & 20\% & 5.00 & 100.00 & 20.00 & 4.98 & 99.60 & 19.92 & 4.98 & 99.60 & 19.92 \\
 Naturalness              & 20\% & 4.90 & 98.00  & 19.60 & 4.43 & 88.60 & 17.72 & 4.41 & 88.20 & 17.64 \\
\cmidrule(lr){1-11}
\textit{Total}            & 100\% &      &        & \textbf{99.00} &     &       & \textbf{95.70} &     &       & \textbf{95.40} \\
\bottomrule
\end{tabular}
}

\end{table}

Table~\ref{tab:app:task_accuracy_full} reports the full per-sub-task rubric behind Table~\ref{tab:exp:task_accuracy}. Each service's score is a weighted aggregate of sub-task scores; for each sub-task we report the raw judge score, its relative score against the per-sub-task reference, and its weighted contribution. Scores are averaged over a panel of three frontier LLM judges under cross-evaluation to reduce single-judge and self-preference bias. Conditional sub-tasks (task suggestion) fire on only a subset of the 100 evaluation calls, which is why their scores are noisier than sub-tasks evaluated on every call.

\clearpage
\section{On-Device Performance: Speed and Model Size}
\label{sec:appendix:ondevice_perf}
\begin{table}[H]
\centering
\caption{Prefill and decode throughput (token/sec) and model size (MiB) across OS, Hardware, and Frameworks. Dashes mark unsupported configurations. All models were benchmarked with Max 4K Tokens (3K prompt tokens and 0.5$\sim$1K generated tokens)}
\label{tab:app:ondevice_perf_full}
\vspace{1em}
% \scriptsize
% \setlength{\tabcolsep}{1pt}
% \renewcommand{\arraystretch}{1.05}
\resizebox{\textwidth}{!}{
\begin{tabular}{l c l rrr rrr}
\toprule
& & & \multicolumn{3}{c}{\textbf{LLaMA3.2 1B}}
    & \multicolumn{3}{c}{\textbf{EXAONE4.0 1.2B}} \\
\cmidrule(lr){4-6}\cmidrule(lr){7-9}
\textbf{OS} & \textbf{Hardware} & \textbf{Framework}
& \textbf{Prefill} & \textbf{Decode} & \textbf{Size}
& \textbf{Prefill} & \textbf{Decode} & \textbf{Size} \\
\midrule
\multicolumn{9}{l}{\textit{Test Device 1: Samsung Galaxy S25 Ultra (Snapdragon 8 Elite)}} \\
\midrule
Android & CPU & llama.cpp & 195.0&  40.0 & 728\,MiB &  49.4 & 39.4 & 786\,MiB \\
        &     & ExecuTorch & 526.4&  66.3 & 821\,MiB & 145.7 & 54.2 & 853\,MiB \\
        % &     & Ours      &   Z.bc &   X.bc &  YX.ac &   Z.bc &   X.bc &  YX.ac \\
\addlinespace[2pt]
\cdashline{1-9}[3.5pt/2.5pt]
\addlinespace[2pt]
% \midrule
        & NPU  & llama.cpp &  132.4 & 25.5 &   851\,MiB &   97.2 & 17.1 &   887\,MiB \\
        &      & ExecuTorch & 2639.5 & 42.6 & 1,037\,MiB & 2327.1 & 32.5 & 1,026\,MiB \\
        &      & \cellcolor{gray!12}Ours & \cellcolor{gray!12}5025.1 & \cellcolor{gray!12}64.7 &  \cellcolor{gray!12}980\,MiB &   \cellcolor{gray!12}3558.2 &  \cellcolor{gray!12}55.8 &  \cellcolor{gray!12}803\,MiB \\
\midrule
\multicolumn{9}{l}{\textit{Test Device 2: Apple iPhone 17 Pro (A19 Pro)}} \\
\midrule
iOS     & CPU & llama.cpp & 356.2 &  24.7 &  763\,MiB & - & - & - \\
        &     & ExecuTorch & 160.3 &  25.7 &  822\,MiB & - & - & - \\
%       &     & Ours      &   Z.bc &   X.bc &  YX.ac &   Z.bc &   X.bc &  YX.ac \\
\addlinespace[2pt]
\cdashline{1-9}[3.5pt/2.5pt]
\addlinespace[2pt]
% \midrule
        & NPU & llama.cpp & - & - & - & - & - & - \\
        &     & ExecuTorch & - & - & - & - & - & - \\
        &     & ANEMLL & 418.2 &  27.9 & 729\,MiB & - &  - &   - \\
        &     & \cellcolor{gray!12}Ours      &   \cellcolor{gray!12}485.1 &   \cellcolor{gray!12}32.1 &  \cellcolor{gray!12}735\,MiB &   \cellcolor{gray!12}412.2 &   \cellcolor{gray!12}19.5 &  \cellcolor{gray!12}783\,MiB \\
\bottomrule
\end{tabular}
}
\end{table}

Table~\ref{tab:app:ondevice_perf_full} reports prefill/decode throughput (tokens/s) and model size for EXAONE4.0 1.2B and LLaMA3.2 1B on the Samsung Galaxy S25 Ultra (Qualcomm Snapdragon 8 Elite) and Apple iPhone 17 Pro (A19 Pro), across CPU and NPU and across ExecuTorch, llama.cpp, and our platform. For frameworks other than ours, models were converted using the stable conversion options provided by each open-source framework and run with the default inference options of the corresponding runtime.
On the Snapdragon NPU ixi-sLM occupies 803 MiB against ExecuTorch's 1{,}026\,MiB, a 22\% reduction combining vocabulary trimming with
module-wise mixed precision.

\clearpage
\section{On-Device Performance: Power and Thermal}
\label{sec:appendix:power_thermal}

\subsection{Power}
\label{sec:appendix:power}

\begin{figure}[H]
\centering
\includegraphics[width=0.9\textwidth]{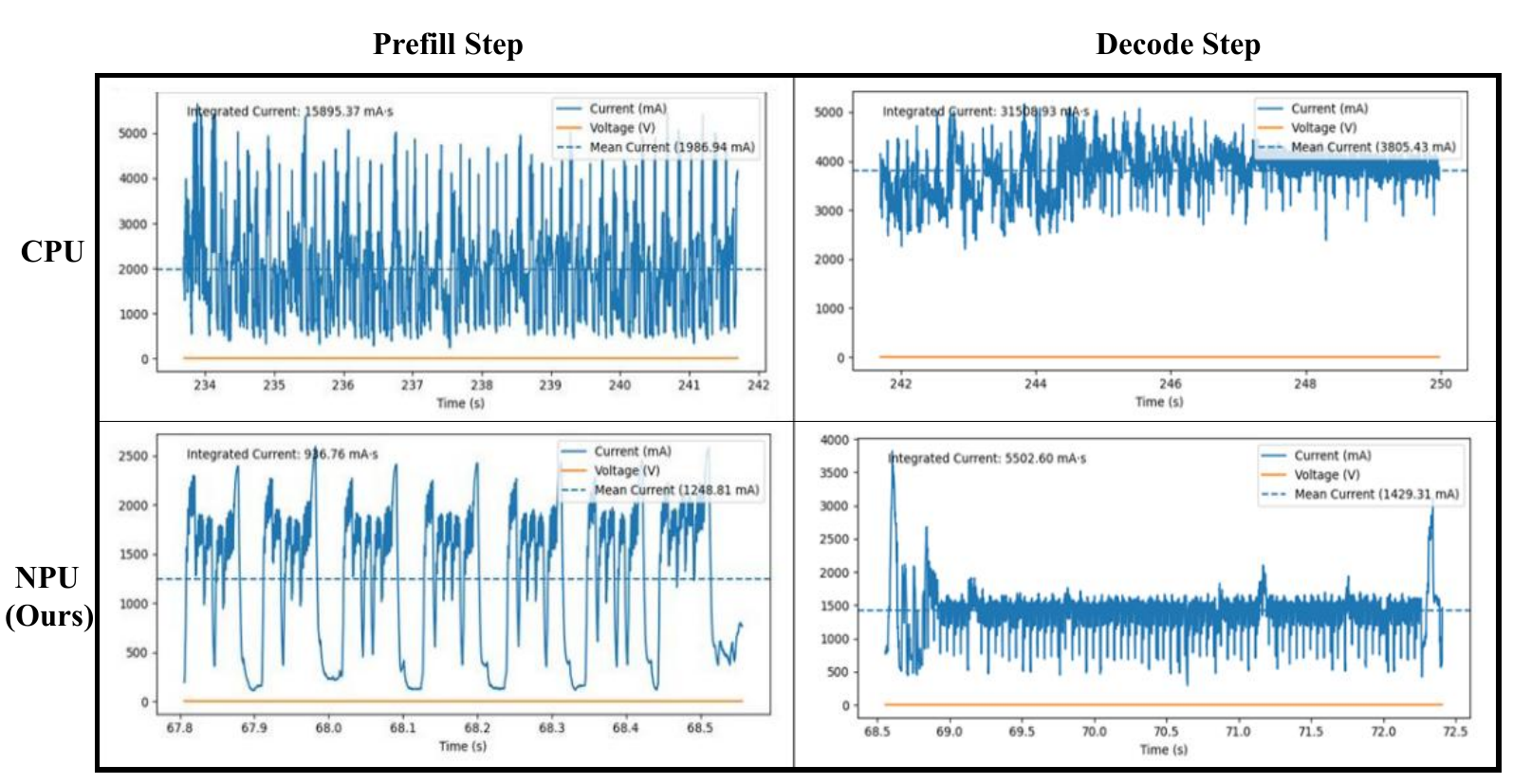}
\caption{Battery current traces over one representative call, CPU vs.\ NPU, segmented into prefill and decode phases.}
\label{fig:app:power}
\end{figure}

\begin{wraptable}{r}{0.6\linewidth}
\centering
\caption{
Per-call Energy (mAh) by phase, CPU vs.\ NPU.
}
\label{tab:app:ondevice_perf_power}
% \vspace{1em}
\begin{tabular}{l ccc}
\toprule
\textbf{Stage} & \textbf{Prefill} & \textbf{Decode} & \textbf{Total} \\
\midrule
CPU & 4.42 & 3.37 & 7.79 \\
NPU & 0.26 & 1.53 & 1.79 \\
\bottomrule
\end{tabular}
\end{wraptable}
We measure on-device power with a power monitor (Monsoon Power Monitor) sampling battery current at 5\,kHz against a 5{,}000\,mAh, $\sim$4.2\,V reference. The target model (ixi-sLM) serves one representative call ($\sim$4K tokens), segmented into prefill and decode phases. Table~\ref{tab:app:ondevice_perf_power} breaks the per-call energy of Table~\ref{tab:exp:ondevice_perf_summary} down by stage: the NPU's saving is concentrated in the compute-bound prefill phase ($\sim$17$\times$), with a $\sim$2$\times$ saving in the memory-bound decode phase.

\clearpage
\subsection{Thermal}
\label{sec:appendix:thermal}

Sustained inference must also respect the device's thermal envelope: a call agent that throttles mid-call fails its latency contract. Table~\ref{tab:app:ondevice_perf_temperature} reports device temperature at 30\,s intervals over a 300\,s sustained run. 

The CPU crosses the 36\,\textdegree C throttling threshold within the first 60\,s, climbs to ∼45\,\textdegree C, and plateaus there for the remainder of the run, while the NPU peaks at 34.5\,\textdegree C and stays under the threshold throughout, leaving thermal headroom for the rest of the system during an active call. 

\begin{table}[H]
\centering
\caption{
Temperature (\textdegree C) at 30\,s intervals over a 300\,s sustained run, CPU against NPU. The final column is the average.
}
\label{tab:app:ondevice_perf_temperature}
\vspace{1em}
\resizebox{\textwidth}{!}{
\begin{tabular}{l cccccccccccc}
\toprule
\textbf{} & \textbf{0} & \textbf{30} & \textbf{60} & \textbf{90} & \textbf{120} & \textbf{150} & \textbf{180} & \textbf{210} & \textbf{240} & \textbf{270} & \textbf{300} & \textbf{Average}\\
\midrule
CPU & 29.9 & 36.5 & 43.0 & 43.5 & 44.0 & 44.5 & 45.0 & 44.5 & 44.0 & 44.6 & 45.3 & 42.3 \\
NPU (ixi-sLM) & 29.9 & 30.4 & 31.6 & 32.2 & 33.0 & 33.1 & 33.3 & 33.5 & 33.5 & 34.0 & 34.5 & 32.6 \\
\bottomrule
\end{tabular}
}
\end{table}

\begin{figure}[H]
\centering
\includegraphics[width=0.85\textwidth]{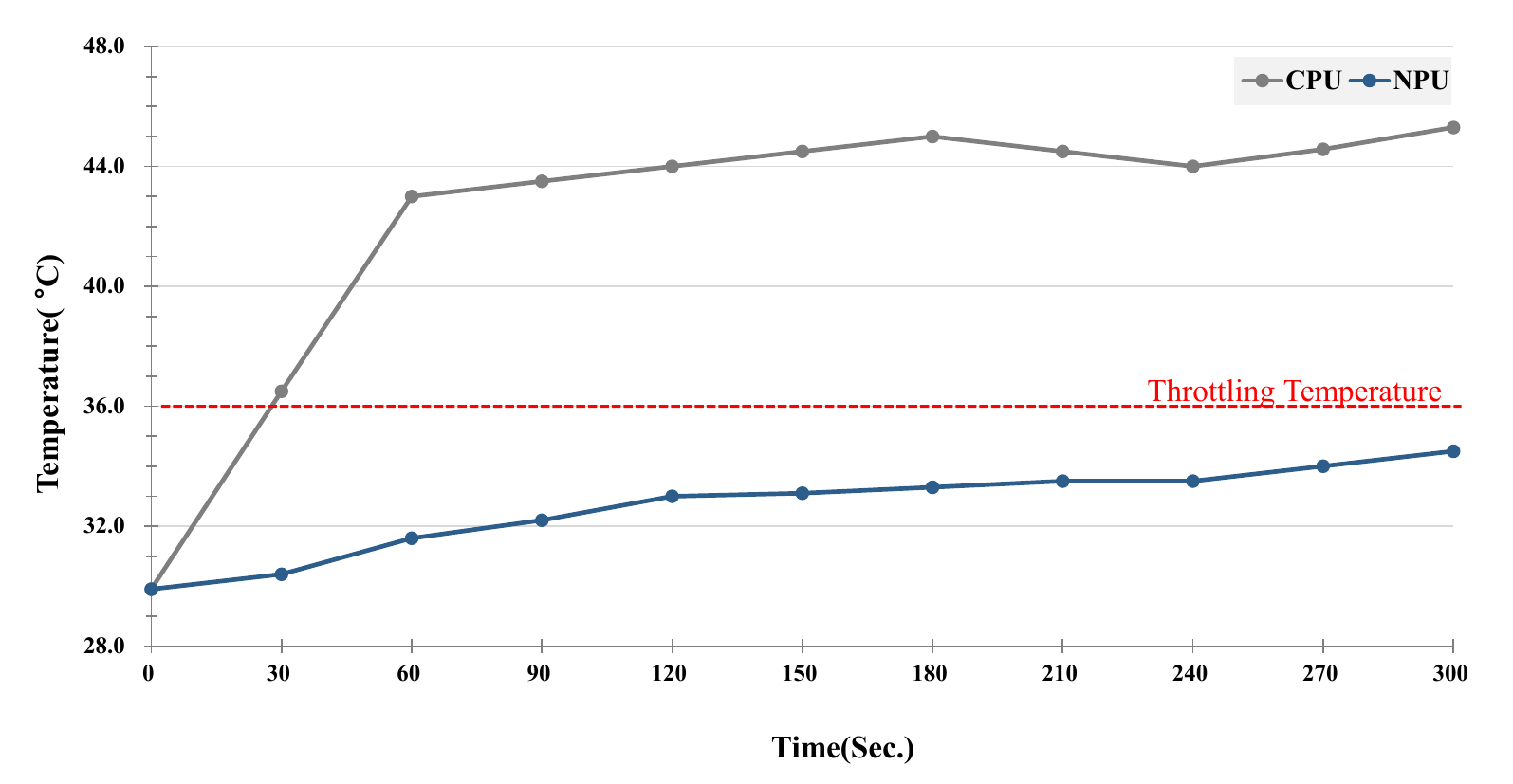}
\caption{Device surface temperature over a 300\,s sustained inference run, CPU vs.\ NPU. The dashed line marks the 36\,\textdegree C throttling threshold.}
\label{fig:app:temperature}
\end{figure}

\clearpage
\section{ixi-O Service Overview}
\label{sec:appendix:ixio_service}

\textbf{AI Call Summary} runs after a call ends: it condenses the conversation into a one-line and a detailed summary with keywords, and proposes follow-up tasks such as calendar events, reminders, and contacts.

\textbf{AI Call Answering} runs when the user is away: the model answers on the user's behalf and converses with the caller in real time while the user follows the exchange.

\begin{figure}[H]
\centering
\includegraphics[width=\textwidth]{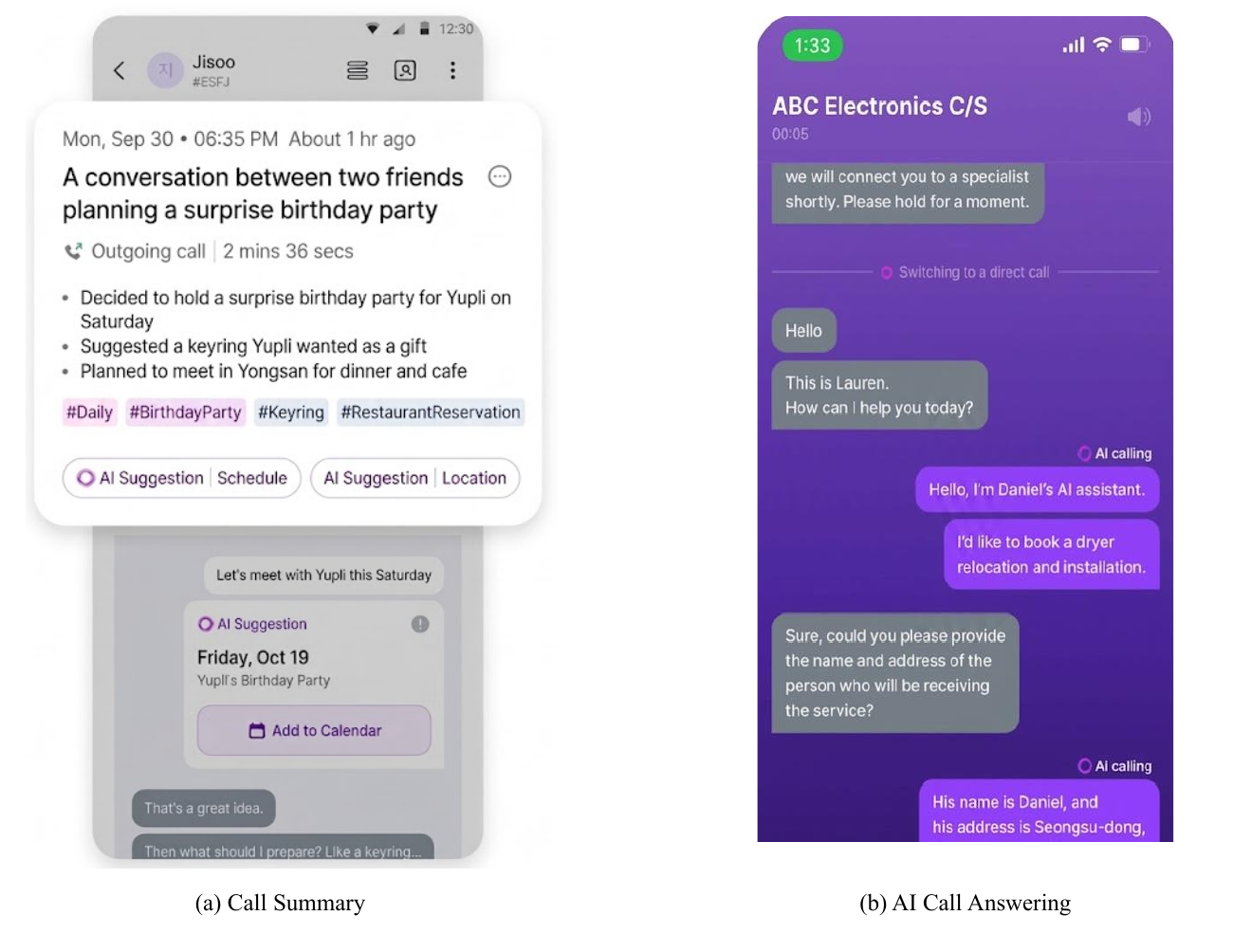}
\caption{\textbf{(a) AI Call Summary:} After a call ends, the agent produces an on-device summary capturing the conversation's intent and key points, classifies it with semantic tags, and surfaces actionable suggestions (e.g., adding an event to the calendar). \textbf{(b) AI Call Answering:} The agent autonomously carries out a voice conversation on behalf of the user, handling task-oriented dialogue with an external party to fulfill the user's request.}
\label{fig:app:ixioui}
\end{figure}

Figure~\ref{fig:app:ixioui} shows both experiences; the sub-tasks evaluated in the task-accuracy rubric (Appendix~\ref{sec:appendix:acc}) correspond directly to these service functions.

\clearpage
\section{Service Prompt}
\label{sec:appendix:ixio_prompt}

This appendix shows the input and output prompt format used by ixi-sLM. The example below is a synthetic illustration created for exposition only; it was not used for model training, calibration, evaluation, or production inference.

\vspace{0.5em}
\noindent
\begin{minipage}[t]{0.48\textwidth}
\begin{safeex}[title=(Korean) Call Summarization Input Prompt, left=5pt, right=5pt, top=4pt, bottom=4pt, boxsep=1.5pt, before skip=0pt, after skip=0pt]
\scriptsize
\textbf{Call Time}\\
2025-04-18 10:30:32

\vspace{0.2em}
\textbf{Call Transcript}

\textbf{TX}: 안녕하세요 상무님 논의 사항 있어서 전화드렸습니다. 나노기술 연구에서 최근 어떤 진전이 있었나요.\\
\textbf{RX}: 네 부팀장님도 안녕하세요. 최근에 나노소재를 이용한 신약 개발에 큰 진전이 있었습니다.\\
\textbf{TX}: 그 신약 개발 프로젝트에 대해 좀 더 자세히 설명해 주시겠어요.\\
\textbf{RX}: 네. 나노입자를 이용해 약물 전달 효율을 크게 향상시켰습니다.\\
\textbf{TX}: 그럼 이 기술이 상용화되기까지 얼마나 걸릴까요.\\
\textbf{RX}: 현재 임상 시험 중이며 상용화까지는 약 2년 정도 예상됩니다.\\
\textbf{TX}: 임상 시험 결과는 어떤가요 긍정적인가요.\\
\textbf{RX}: 네 지금까지의 결과는 매우 긍정적입니다. 부작용도 거의 없습니다.\\
...
\end{safeex}
\end{minipage}
\hfill
\begin{minipage}[t]{0.48\textwidth}
\begin{artifactex}[title=(Korean) Call Summarization Output, left=5pt, right=5pt, top=4pt, bottom=4pt, boxsep=1.5pt, before skip=0pt, after skip=0pt]
\scriptsize

\textbf{한줄요약}\\
나노 기술 개발 관련 논의 완료

\vspace{0.35em}\hrule\vspace{0.35em}

\textbf{상세요약} \\
나노 기술 연구 진행 상황에 대해 문의함.\\
나노 소재를 이용한 신약 개발 진전 상황을 설명함.\\
신약 상용화까지 2년 예상되며 임상 시험 결과는 매우 긍정적이라고 보고함.\\
...

\vspace{0.35em}\hrule\vspace{0.35em}

\textbf{키워드} \\
\#업무, \#회의, \#나노기술

\vspace{0.35em}\hrule\vspace{0.35em}

\textbf{AI Task Suggestion}\\
{\ttfamily task: schedule\\
task\_title: 프로젝트 회의\\
date: 2025-04-22\\
time: 14:00:00\\
reference: 20}\\
...

\end{artifactex}
\end{minipage}

\par\vspace{1.0em}
\noindent
\begin{minipage}[t]{0.48\textwidth}
\begin{safeex}[title=(English) Call Summarization Input Prompt, left=5pt, right=5pt, top=4pt, bottom=4pt, boxsep=1.5pt, before skip=0pt, after skip=0pt]
\scriptsize
\textbf{Call Time}\\
2025-04-18 10:30:32

\vspace{0.2em}
\textbf{Call Transcript}

\textbf{TX}: Hello, Director. I called because there are some matters to discuss. What recent progress has been made in nanotechnology research?\\
\textbf{RX}: Hello, Deputy Team Lead. Recently, there has been significant progress in developing new drugs using nanomaterials.\\
\textbf{TX}: Could you explain that new drug development project in more detail?\\
\textbf{RX}: Yes. We have greatly improved drug-delivery efficiency using nanoparticles.\\
\textbf{TX}: Then how long will it take for this technology to be commercialized?\\
\textbf{RX}: It is currently in clinical trials, and commercialization is expected to take about two years.\\
\textbf{TX}: How are the clinical trial results? Are they positive?\\
\textbf{RX}: Yes, the results so far are very positive. There have been almost no side effects.\\
...
\end{safeex}
\end{minipage}
\hfill
\begin{minipage}[t]{0.48\textwidth}
\begin{artifactex}[title=(English) Call Summarization Output, left=5pt, right=5pt, top=4pt, bottom=4pt, boxsep=1.5pt, before skip=0pt, after skip=0pt]
\scriptsize

\textbf{One-line Summary}\\
Discussion on nanotechnology development completed

\vspace{0.35em}\hrule\vspace{0.35em}

\textbf{Detailed Summary} \\
Asked about the progress of nanotechnology research.\\
Explained progress in new drug development using nanomaterials.\\
Reported that commercialization is expected in two years and that clinical trial results are very positive.\\
...

\vspace{0.35em}\hrule\vspace{0.35em}

\textbf{Keywords} \\
\#work, \#meeting, \#nanotechnology

\vspace{0.35em}\hrule\vspace{0.35em}

\textbf{AI Task Suggestion}\\
{\ttfamily task: schedule\\
task\_title: Project meeting\\
date: 2025-04-22\\
time: 14:00:00\\
reference: 20}\\
...

\end{artifactex}
\end{minipage}

\clearpage
\section{Limitations}
\label{sec:appendix:limitations}

Our task-accuracy scores rely on an LLM judge panel, and because one judge belongs to the same model family as the cloud baseline, a residual self-preference may inflate that baseline's scores; cross-evaluation and averaging only partially offset it. The evaluation is also narrow in scope: two services in a single Korean call domain, scored on 100 held-out calls, so the absolute numbers should be read as evidence for this production deployment rather than as a general benchmark. Our results further hold at one operating point, a 1.2B model under a $\sim$1\,GiB budget on two current NPUs; generalization across hardware is an active engineering effort, and the same recipe is now being extended to eight chipsets. Finally, quality is hardest to measure on conditional sub-tasks such as task suggestion, which are scored on few calls and are therefore sensitive to case frequency rather than to a uniform quantization effect; larger, balanced samples are needed to establish their true on-device quality.

% \section{Technical appendices and supplementary material}
% Technical appendices with additional results, figures, graphs, and proofs may be submitted with the paper submission before the full submission deadline (see above). You can upload a ZIP file for videos or code, but do not upload a separate PDF file for the appendix. There is no page limit for the technical appendices. 

% Note: Think of the appendix as ``optional reading'' for reviewers. The paper must be able to stand alone without the appendix; for example, adding critical experiments that support the main claims to an appendix is inappropriate. 

%%%%%%%%%%%%%%%%%%%%%%%%%%%%%%%%%%%%%%%%%%%%%%%%%%%%%%%%%%%%

%\newpage
%\input{checklist.tex}

\end{document}